\documentclass{article}
\usepackage{spconf,amsmath,graphicx}
\usepackage{url}

\title{A GENERIC ENSEMBLE BASED DEEP CONVOLUTIONAL NEURAL NETWORK FOR SEMI-SUPERVISED MEDICAL IMAGE SEGMENTATION}

\name{Ruizhe Li$^{1}$ \qquad Dorothee Auer$^{2}$ \qquad Christian Wagner$^{1}$ \qquad Xin Chen$^{1}$}
\address{$^{1}$IMA/LUCID Group, School of Computer Science, University of Nottingham, UK\\
        $^{2}$School of Medicine, University of Nottingham, UK}
\begin{document}
%
\maketitle
\begin{abstract}
Deep learning based image segmentation has achieved the state-of-the-art performance in many medical applications such as lesion quantification, organ detection, etc. However, most of the methods rely on supervised learning, which require a large set of high-quality labeled data. Data annotation is generally an extremely time-consuming process. To address this problem, we propose a generic semi-supervised learning framework for image segmentation based on a deep convolutional neural network (DCNN). An encoder-decoder based DCNN is initially trained using a few annotated training samples. This initially trained model is then copied into sub-models and improved iteratively using random subsets of unlabeled data with pseudo labels generated from models trained in the previous iteration. The number of sub-models is gradually decreased to one in the final iteration. We evaluate the proposed method on a public grand-challenge dataset for skin lesion segmentation. Our method is able to significantly improve beyond fully supervised model learning by incorporating unlabeled data. The code is available on Github \footnote{https://github.com/ruizhe-l/semi-segmentation}.
\end{abstract}
\section{INTRODUCTION}
\label{sec:intro}

\begin{figure*}
    \centering
    \includegraphics[width=14cm]{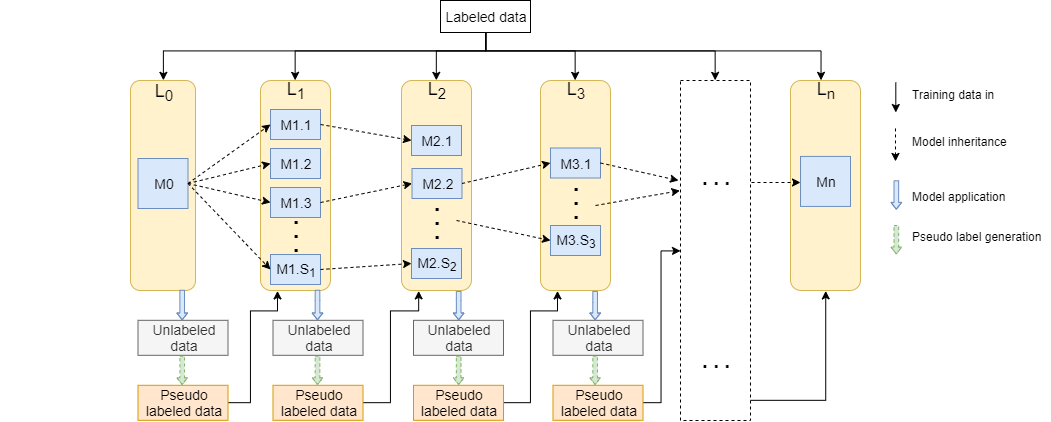}
    \vspace{-0.4cm}
    \caption{Overview of the proposed framework.}
    \label{fig:1}
    \vspace{-0.6cm}
\end{figure*}

Image semantic segmentation is a basic and crucial step for many biomedical image analysis tasks (e.g. tumour quantification). Nowadays, many encoder-decoder based deep convolutional neural networks (DCNNs) such as U-Net \cite{ronneberger2015u} achieve state-of-the-art performance for image segmentation using fully-supervised learning. However, data annotation is extremely time consuming especially for medical imaging where highly skilled expertise is required.

Several methods have been proposed to address this challenge. Data augmentation is commonly used as an effective solution. A few studies show that geometric transformations and intensity shifts to increase the number of annotated data can achieve better performance than only using the original labeled data \cite{cui2015data}. Moreover, semi-supervised learning methods that use both labeled and unlabeled data for model training have also been intensively studied. In 2015, Rasmus et al. \cite{rasmus2015semi} adapted the Ladder Network for simultaneously minimizing the sum of supervised and unsupervised cost functions by backpropagation for image classification. Baur et al. \cite{baur2017semi} proposed a semi-supervised learning method based on fully convolutional networks (FCNs) and random feature embedding for image segmentation. More recently, Lahiri et al. \cite{lahiri2018retinal} used the discriminator in generative adversarial networks (GANs)  to both segment images and discriminate between real and generated fake images. Labeled images help to improve the segmentation accuracy, and unlabeled images are used to increase the discrimination power. However, this method only extracts the global information of images to improve the ability to discriminate true and false samples but does not pay attention to the extraction of segmentation information. More closely related to our proposed method is \cite{bai2017semi}. They developed a network-based semi-supervised learning framework using self-learning techniques. A fully supervised model is firstly trained on labeled data, then pseudo labels are generated for unlabeled data using this model and refined by a fully-connected conditional random field (CRF). Subsequently, both labeled data and pseudo labeled data are used to refine the initial model. This process is repeated until convergence. However, the quality of the pseudo labels is highly dependent on the initial model and has a high impact on the final performance. 

Similar to \cite{bai2017semi}, in this paper, we also train an initial model using labeled data and subsequently refine the model using unlabeled data with pseudo labels. Different from \cite{bai2017semi}, we use an ensemble technique to reduce the negative influence of poor quality pseudo labels. Ensemble learning is a widely used technique in machine learning \cite{dietterich2000ensemble}. It trains multiple weak classifiers and then combines these “weak decisions” to generate a final solution. Several classical classifiers (e.g. random forests \cite{niemeyer2013classification}) are based on this idea to achieve state-of-the-art performance prior to the arrival of deep learning methods. Due to heavy computational load, very few studies have integrated ensemble learning with deep learning models. In 2017, Dolz et al \cite{dolz2017deep} proposed a suggestive annotation model for infant brain MRI segmentation in a fully supervised manner. It achieves the state-of-the-art performance by combining 10 CNNs. To the best of our knowledge, no previous studies have reported the use of ensemble learning and DCNN to achieve semi-supervised learning.

In this paper, we propose a DCNN based semi-supervised learning framework that aims to learn a generic model that only uses a few annotated data with some unlabeled data for model training. We evaluate our framework based on a public skin lesion segmentation dataset, showing it outperforms both fully supervised learning method using only labeled data and Bai's method \cite{bai2017semi} by a large margin.

\section{METHODOLOGY}
\label{sec:method}

An overview of the framework is illustrated in \textbf{Fig. \ref{fig:1}}. The proposed framework is an iterative process, where the indices of iterations are represented by \{$L_0$, …, $L_n$\}. A well-established encoder-decoder DCNN network (U-net \cite{ronneberger2015u}) is used as the basic segmentation method to train an initial model ($M0$) in $L_0$ using all labeled data. Subsequent models from $L_1$ to $L_n$ are trained based on both labeled data and a total number of $N_0$ unlabeled data with pseudo labels generated from the previous iteration. The pseudo labels are generated using a weighted combination of outputs from all sub-models. The parameters of sub-models ($M1.1$, …, $M1.S_1$) in $L_1$ are copied from $M0$. From $L_1$ onwards, the number of sub-models is reduced at each iteration and becomes one in the final iteration ($L_n$). Sub-models in the current iteration are randomly inherited from the sub-models of the previous iteration. The number of unlabeled images (\{$N_0$, …, $N_n$\}) used for training of each sub-model is gradually increased from level to level, following the rule of $N_n = \lfloor N_0/S_n + 0.5 \rfloor$ where $\lfloor \cdot \rfloor$ indicates the mathematical floor operation. Finally, $M_n$ trained on all labeled data and unlabeled data with pseudo labels is the final model used for segmenting any unseen data.

\subsection{Initial Supervised Segmentation Model}
\label{ssec:method-m0}

U-net proposed by Ronnebergeret et al. \cite{ronneberger2015u} is a DCNN based encoder-decoder network which has achieved state-of-the-art performance for many image segmentation tasks in medical applications. U-net has an encoder and a decoder that both consist of several layers of feature maps by applying two $3\times3$ convolutional operations and one rectified linear unit (relu) at each layer. In the encoding path, max-pooling with a stride of 2 is performed between two consecutive layers to achieve feature map down-sampling. Symmetrically, the decoder uses up-convolution to up-sample the feature map from the previous layer. Additionally, there are skip paths that concatenate the feature maps from the encoder to the corresponding layers of the decoder. A $1\times1$ convolution is used in the final decoder layer to convert the dimension of feature map to the number of classes. Subsequently, the softmax function is applied to map the activation values at each pixel position to the range of [0, 1]. The cost function we used in our work combines both cross-entropy and Dice coefficient with equal weights, which outperforms the sole use of cross-entropy and was suggested by Liu et al. \cite{liu2019bladder}. As a commonly used improvement, we also add the residual block \cite{he2016deep} to the conventional block in the U-net for faster convergence. A validation dataset is used for determining the termination point of model training. Detailed parameter settings are provided in \textbf{Section \ref{ssec:experiments-params}}.

\subsection{Model Improvements using Unlabeled Data}
\label{ssec:method-submodels}

Based on the initial segmentation model, we further improve the model using unlabeled data. We firstly apply model $M0$ to all $N_0$ unlabeled data to generate segmentation outputs (probability of each pixel belonging to each class label) which serve as pseudo labels. Then a random sub-set of these unlabeled data with pseudo labels, together with the labeled data, are used to train a number of sub-models ($M1.1$,…,$M1.S_1$), which are called level one models in iteration L1. The initial parameters of these sub-models are the same and copied from $M0$. We initially generate $S_1$ sub-models, each with $N_1$ training data. The validation dataset for $M0$ is also used in subsequent levels to prevent the sub-models from model overfitting. For sub-model training from $L_2$ to $L_n$, the number of sub-models ($S_n$) is reduced as follows: 
\vspace{-0.2cm}
\begin{equation}
    \label{eq:1}
    \vspace{-0.2cm}
    S_n = max(\lfloor \frac{S_1}{2^{n-1}} \rfloor, 1), \ for \ n > 1
\end{equation}

The whole framework stops training when $S_n$ reaches 1. The training image contains a number of unlabeled data and all labeled data. Hence the effect of the labeled data is gradually reduced when the number of unlabeled images increases from level to level. This mechanism enables the model to gradually learn more information from unlabeled data without a sudden performance drop. 

From $L_2$ to $L_n$, the number of sub-models is reduced. They are randomly selected from the sub-models in previous level. When all models in a level have finished their training, a new set of pseudo labels ($P_k$, $k=1$,…, $N_0$) are generated for all the unlabeled data based on a weighted combination of the output pixel-wise probability map ($M_{i,k}$) from all sub-models using:
\begin{equation}
    \label{eq:2}
    P_k = \sum_{i=1}^{S_n}w_i M_{i,k}
\end{equation}
The weight $w_i$  for each sub-model is calculated as follows:
\begin{equation}
    \label{eq:3}
    w_i = \sum_{j=1}^{R}B_{i,k}^j C_k^j
\end{equation}
where $B_{i,k}^j$ is the binary map by applying a threshold of 0.5 to the pixel-wise probability map generated from the $i^{th}$ sub-model for the $k^{th}$ unlabeled data. Superscript $j$ is the index of pixels in an image that contains a total of $R$ pixels. $C_k^j$ is the summation of all pixel-wise probability maps generated from all sub-models for the $k^{th}$ unlabeled image. Intuitively, the weight of the $i^{th}$ sub-model is calculated as the sum of probability values of the combined outputs from all sub-models within the image region predicted by the $i^{th}$ sub-model. A larger weight indicates a greater agreement between the individual prediction of a sub-model and the combined prediction of all sub-models. 
We scale the weights to the range of [0.1, 1] using Eq. (\ref{eq:4}), maximising the distance between the performance of the best and worst sub-models and thus introducing effectively a `relative reward' -- rewarding the best sub-model the most, the worst sub-model the least, and applying a relative distribution between them. 
\begin{equation}
    \label{eq:4}
    w_i = \frac{w_i-min(w)}{max(w)-min(w)} \times 0.9 + 0.1
\end{equation}
Finally, weights are also normalised by dividing the sum of all weights. These new pseudo labels are used to train the sub models in the next level.

The task of sub-models is to learn features that are potentially different from the initial supervised model. The outputs of these sub-models are then regulated and aggregated by the weighted combination process. These two processes work interactively to improve the segmentation performance. During this iterative process from level $L_2$ to $L_n$, the number of sub-models is decreased and the number of training images per sub-model is increased. The whole framework will be terminated at level $L_n$ when only one model is trained using all labeled and unlabeled data. 

\section{EXPERIMENTS AND RESULTS}
\label{sec:experiments}

\begin{table*}
    
    \centering
    \caption{Comparison of our proposed method to the FS-100 and FS-2044 models and Bai's method. Mean $\pm$ standard deviation values are reported.}
    \label{table:1}
    \vspace{0.1cm}
    \begin{tabular}{|l|l|l|l|l|l|l|}
        \hline
        \textbf{Method} & \textbf{DC}   & \textbf{IoU}  & \textbf{Accuracy} & \textbf{Sensitivity} & \textbf{Specificity} & \textbf{Training time (s)} \\ \hline
        FS-2044       & $0.872\pm0.137$ & $0.793\pm0.171$ & $0.952\pm0.072$ & $0.882\pm0.156$ & $0.974\pm0.052$ & $5926$ \\ \hline
        FS-100        & $0.793\pm0.190$ & $0.693\pm0.218$ & $0.924\pm0.103$ & $0.859\pm0.180$ & $0.956\pm0.076$ & $2314$ \\ \hline
        Bai’s         & $0.817\pm0.189$ & $0.724\pm0.215$ & $0.933\pm0.103$ & $0.827\pm0.203$ & $0.970\pm0.073$ & $18822$ \\ \hline
        Proposed      & $0.844\pm0.156$ & $0.755\pm0.189$ & $0.939\pm0.091$ & $0.887\pm0.159$ & $0.969\pm0.062$ & $27853$ \\ \hline
    \end{tabular}
    \vspace{-0.3cm}
\end{table*}

\def \hgap {-0.4cm}
\def \vgap {-0.1cm}
\def \imgsize {1.8cm}

\begin{figure*}[th]
    \centering
    \begin{tabular}{cccccc}
        \textbf{a} &
        \hspace{\hgap} \textbf{b} &
        \hspace{\hgap} \textbf{c} &
        \hspace{\hgap} \textbf{d} &
        \hspace{\hgap} \textbf{e} &
        \hspace{\hgap} \textbf{f} \\
        \vspace{\vgap}
        \includegraphics[width=\imgsize]{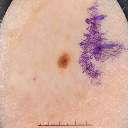} &
        \hspace{\hgap} \includegraphics[width=\imgsize]{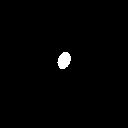} &
        \hspace{\hgap} \includegraphics[width=\imgsize]{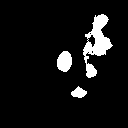} &
        \hspace{\hgap} \includegraphics[width=\imgsize]{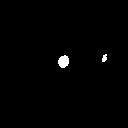} &
        \hspace{\hgap} \includegraphics[width=\imgsize]{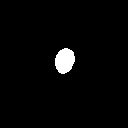} &
        \hspace{\hgap} \includegraphics[width=\imgsize]{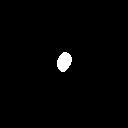} \\
        \vspace{\vgap}
        \includegraphics[width=\imgsize]{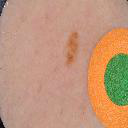} &
        \hspace{\hgap} \includegraphics[width=\imgsize]{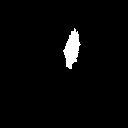} &
        \hspace{\hgap} \includegraphics[width=\imgsize]{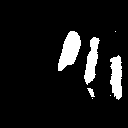} &
        \hspace{\hgap} \includegraphics[width=\imgsize]{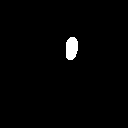} &
        \hspace{\hgap} \includegraphics[width=\imgsize]{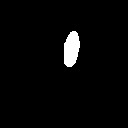} &
        \hspace{\hgap} \includegraphics[width=\imgsize]{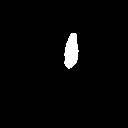} \\
        \vspace{\vgap}
        Input & 
        \hspace{\hgap} GT & 
        \hspace{\hgap} FS-100 & 
        \hspace{\hgap} Bai’s& 
        \hspace{\hgap} Proposed & 
        \hspace{\hgap} FS-2044
    \end{tabular}
    \vspace{-0.2cm}
    \caption{Qualitative assessment for different models.}
    \label{fig:2}
    \vspace{-0.5cm}
\end{figure*}

\subsection{Materials and Experiments}
\label{ssec:experiments-data}

The proposed method was evaluated on “ISIC 2018: Skin Lesion Analysis Towards Melanoma Detection” grand challenge dataset \cite{codella2019skin}\cite{tschandl2018ham10000}.   It provides 2594 images with manually annotated lesions (binary class). We pre-processed the image by zero-mean normalization (i.e. original intensity subtracted by the mean image intensity and divided by standard deviation). The original image size varies, which was resized to $128\times128$ to achieve a balance between computational efficiency and accuracy. We also tested our method with larger image sizes (e.g. $256\times256$), which did not lead to significantly better performance but with much longer computational time, as the shapes of the lesions are not very complicated.

For our method, we split the dataset into a training set and a test set with ratio of approximately 80\%/ 20\% (i.e 2094/ 500 images). Within the training set, we further split the data into a labeled set, an unlabeled set and a validation set that contain 100, 1944 and 50 images respectively. 

For comparison, we also trained fully supervised models using only the 100 labeled images (FS-100), and the full set of 2044 labeled images in the training set (FS-2044). The FS-100 result serves as a baseline to demonstrate the improvement of our method by including unlabeled data. The FS-2044 result serves as the upper bound to indicate the best possible performance using all training images as labeled data. Bai’s self-training method \cite{bai2017semi} as described in \textbf{Section \ref{sec:intro}} was also implemented for comparison.  

\subsection{Parameter Settings}
\label{ssec:experiments-params}

For all methods, the basic network structures were the same, which was U-net with residual block (described in \textbf{Section \ref{ssec:experiments-data}}). It consisted of 5 encoder and decoder layers respectively. The number of root features was 16 and doubled at the next layer in the encoder path and halved in the decoder path. 

For our method, the initial model was trained using 100 images for a maximum of 200 epochs with batch size of 10. The learning rate was 0.0001, and dropout rate was 25\%. Due to the data size for sub-models being dynamic, the batch size we used for sub-models was 1. For sub-model training, the maximum number of epochs was 50 and the learning rate was 0.0001. Ensemble learning is capable to reduce bias and avoid overfitting, therefore the dropout is removed for faster convergence. The training process is stopped early when the loss value of the validation set is not decreased for 5 consecutive epochs. We tested 32, 16 and 8 as the number of sub-models in $L_1$ with 6, 5 and 4 levels respectively. Our results showed that a total number of 5 levels with the number of sub-models 16, 8, 4, 2, and 1 for $L_1$, $L_2$, $L_3$, $L_4$ and $L_5$ achieved the best performance in terms of balancing the computational time and segmentation accuracy.

Both fully-supervised methods FS-100 and FS-2044 were trained for 200 epochs with a learning rate of 0.0001. No early stopping was used but a drop out rate of 0.25 was applied to avoid over-fitting. The batch sizes for FS-100 and FS-2044 were 10 and 28 respectively. 

For Bai's method, we used the same 100 labeled images as in our method to train an initial model and further refined it by including 1944 unlabeled images. Following the paper, we applied fully-connected CRF on all pseudo labels in each iteration. After evaluating the CRF method on the validation set, we set the CRF parameters as $w_1 = 2$, $w_2 = 1$, $\sigma_\alpha = 2$, $\sigma_\beta = 3$, $\sigma_\gamma = 5$ (refer to \cite{bai2017semi} for more details). The self-training optimisation was performed for 3 iterations, with 50 epochs for each iteration. We also experimented with more iterations, but it did not improve the performance further. Same as FS-2044, the batch size of this model was 28.

\subsection{Results}
\label{ssec:experiments-results}

We report quantitative results measured by Dice coefficient (DC), Intersection over Union (IoU), accuracy, sensitivity, specificity and training time for performance comparison, as shown in \textbf{Table \ref{table:1}}.

It is seen from \textbf{Table \ref{table:1}} that the FS-100 model and FS-2044 model produced the worst and best results respectively, as expected. Both Bai’s method and our method achieved better segmentation accuracies than the baseline FS-100 model, indicating successfully incorporating unlabeled data for model improvement. More importantly, our method is significantly ($p<0.001$ based on paired t-test) better than Bai’s  method in terms of DC, IoU and sensitivity measurements. For this dataset, our method produced the best sensitivity values even better than the FS-2044 method but slightly lower specificity caused by more false positives.

Our method requires the training of a number of sub-models that leads to longer training time than Bai’s method. Since the sub-models in earlier iterations (e.g. L1 and L2) only use a few images for training and the model parameters are inherited from previous models, the learning process converges rapidly. Hence the training time for the whole process is not tremendously high, particularly compared with a time-consuming manual annotation process.  

For qualitative assessment, some example images are shown in \textbf{Fig. \ref{fig:2}}. \textbf{Fig. \ref{fig:2} (a)} and \textbf{(b)} are the input skin lesion images and their corresponding manual annotations respectively. \textbf{Fig.  \ref{fig:2} (c)-(f)} are the segmentation results from the FS-100 model, Bai’s method, our method and FS-2044 model respectively. Similar to the conclusion drawn from the quantitative results, both Bai’s method and our method greatly improved the segmentation accuracy compared with FS-100 method by reducing false positives significantly. Compared with the ground truth, Bai’s method excessively reduced the size of the target regions (more false negatives), where as our method detected slightly larger region (more false positives). 

\section{DISCUSSION  AND CONCLUSIONS }
\label{conclusions}

By combining ensemble learning and DCNN, we have proposed a generic semi-supervised learning framework that is able to improve fully supervised model by incorporating unlabeled data. By evaluating the method using a public skin lesion dataset, the proposed method has shown a superior performance in comparing with a similar state-of-the-art semi-supervised learning method.

\bibliographystyle{IEEEbib}
\bibliography{strings,refs}

\end{document}